\pdfoutput=1

\documentclass[11pt]{article}

\usepackage[preprint]{acl}

\usepackage{times}
\usepackage{latexsym}
\usepackage[pdftex]{graphicx}
\usepackage[T1]{fontenc}
\usepackage{amssymb}

\usepackage[utf8]{inputenc}

\usepackage{microtype}

\usepackage{inconsolata}
\usepackage{amsfonts}
\usepackage{amsmath}
\usepackage{enumitem}
\newcommand{\shortName}{SynCARS} 
\newcommand{\finalTitle}{Synthesizing Conversations from Unlabeled Documents using Automatic Response Segmentation}
\usepackage{caption}
\usepackage{booktabs}
\usepackage{hyperref}
\DeclareCaptionLabelFormat{AppendixTables}{Table A.#2}
\DeclareCaptionLabelFormat{AppendixFigures}{Figure A.#2}

\title{\finalTitle}

\author{Fanyou Wu \\ Amazon \\ \texttt{fanyouwu@amazon.com}\And
        Weijie Xu \\ Amazon \\ \texttt{weijiexu@amazon.com} \AND 
        Chandan K. Reddy \\ Amazon \\ \texttt{ckreddy@amazon.com}\And
        Srinivasan H. Sengamedu \\ Amazon \\ \texttt{sengamed@amazon.com} 
       }

\begin{document}
\maketitle
\begin{abstract}
In this paper, we tackle the challenge of inadequate and costly training data that has hindered the development of conversational question answering (ConvQA) systems. Enterprises have a large corpus of diverse internal documents. Instead of relying on a searching engine, a more compelling approach for people to comprehend these documents is to create a dialogue system. In this paper, we propose a robust dialog synthesising method called~\shortName. We learn the segmentation of data for the dialog task instead of using segmenting at sentence boundaries.  The synthetic dataset generated by our proposed method achieves superior quality when compared to WikiDialog, as assessed through machine and human evaluations. By employing our inpainted data for ConvQA retrieval system pre-training, we observed a notable improvement in performance across standard benchmark datasets.\footnote{Our model and dataset are publicly available at \url{https://github.com/wufanyou/SynCARS}. }

\end{abstract}

\begin{figure}[t]
  \centering
  \includegraphics[width=\linewidth]{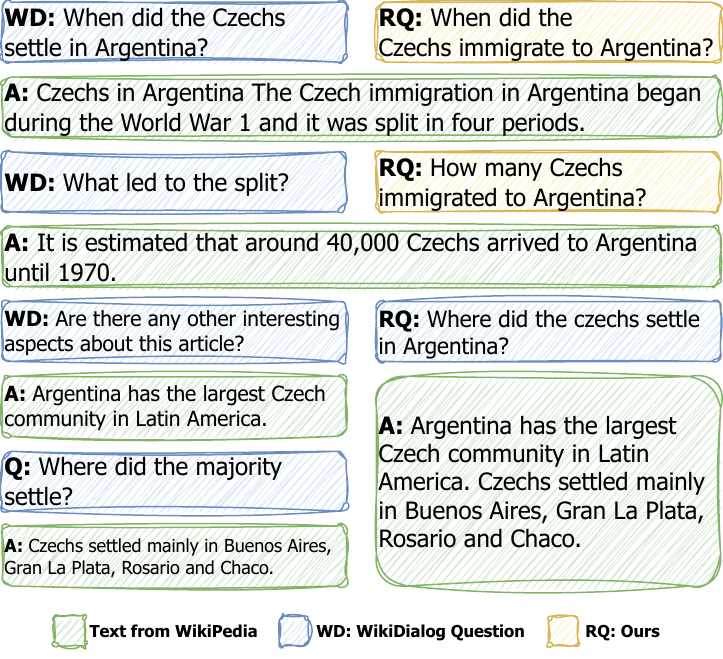}
  \caption{An example dialog from WikiDialog (WD) and ours (RQ). The blue and yellow boxes in the dialog contain the questions generated by our approach, while the green boxes contain the corresponding answers. WD asks a question starting with "Are there any other .." which is not useful to train a question answering system. Besides, some of the answers can be combined such as the last two on the left side. In contrast, our method fixed those problems.}\label{fig:examples}
\end{figure}

\section{Introduction}


Conversational Question Answering (ConvQA) is a computational task aimed at modeling the information-seeking processes found in human dialog. The goal of this task is to allow automated systems to understand and respond to questions within a conversational context. Several publicly available datasets, such as QuAC~\citep{choi2018quac}, CoQA~\citep{reddy2019coqa}, QreCC~\citep{anantha2021qrecc}, and  OR-QuAC~\citep{qu2020open}, have been developed for building ConvQA systems. Despite these resources, the size of the datasets remains relatively limited, posing potential challenges when implementing ConvQA systems in real-world applications.

Meanwhile, a plethora of high-quality documents, including but not limited to sources such as Wikipedia and arXiv, are publicly available. Numerous technological roadmaps have been proposed to leverage the vast wealth of information within these documents to construct a ConvQA system. One approach involves utilizing a Large Language Model (LLM) in conjunction with information retrieval techniques, such as New Bing from Microsoft. Alternatively, a well-trained LLM, such as ChatGPT without a plugin, can be employed independently to achieve similar or even superior results. By utilizing information retrieval tools, LLMs can access up-to-date information, albeit at the expense of increased inference time and latency compared to using LLMs alone.

To effectively employ these documents alongside LLMs for the purpose of constructing a ConvQA system, there are two potential research directions to explore: the decomposition and synthesis of the documents into a dialog format, or the pursuit of improved question embedding representations for the documents. Additionally, conversational question generation (QG) can be utilized in both of these approaches. \citet{dai2022dialog} pioneered "dialog inpainting," suggesting every sentence in a document can answer a question, leading to the creation of the WikiDialog dataset. In our study, we introduce \shortName, a novel approach that leverages ConvQA datasets to produce synthetic conversations from unlabeled documents, yielding superior quality compared to WikiDialog. To summarize, our main contributions can be summarized as follows:
\begin{itemize}[leftmargin=*]

\item We identified various challenges present within the WikiDialog dataset, a component of FLAN collection~\cite{longpre2023flan} used for instruction fine-tuning.

\item We designed a new dataset by merging existing datasets and filtered out inadequate data to address issues present in the WikiDialog dataset. Additionally, we have designed a new answer segmentation technique by introducing a special token $p_{t}^{m}$. In comparison to the approach by~\citet{dai2022dialog}, we obtained a new and compact model specifically tailored  for generating dialogues from documents.

\item Our generated dataset exhibits significantly higher answer quality and question specificity, as validated through Human and \textsc{GPT-4} evaluations, when compared to WikiDialog. Furthermore, the question retrieval system trained on our generated data achieves superior results compared to the system trained on WikiDialog and the standard retrieval-only benchmark method.

\end{itemize}

\section{Related Work}

Question generation (QG) is a field that seeks to create natural questions using various types of data sources, including structured knowledge bases~\citep{guo2018question, chen2023toward}, text~\citep{rus2010first, du2017learning, nogueira2019document}, images~\citep{li2018visual}, and tables~\citep{bao2018table}. Past research efforts in this area have primarily focussed on producing isolated and disconnected questions from a given passage. 

\citet{pan2019reinforced} proposed the Conversational Question Generation (CQG) task as an approach to improve the development of ConvQA systems. This task involves generating the subsequent question by incorporating a passage and a conversation history, thereby requiring a deeper comprehension of the given passage and prior conversation to generate a coherent and relevant question for the next round. Unlike prior QG tasks that only consider the passage, CQG requires an understanding of the previous conversation, making it a more complex task. 

\citet{kim2022generating} proposed SIMSEEK, a framework that generates ConvQA datasets by modeling the information needs of questioners who may ask incoherent questions due to excessive information. SIMSEEK includes a conversational answer extractor that selects answer candidates from the passage by considering the context of the conversation. However, this method is only suitable for short answers.

In contrast, \citet{dai2022dialog} introduced dialog inpainting, which assumes that each sentence in a document can be used as an answer to a question. The authors generated a ConvQA dataset called WikiDialog using this approach.
This dataset tends to have longer answers as each answer corresponds to a single sentence. This characteristic makes it more suitable for dialog applications. While the proposed method is straightforward and efficient, concerns arise regarding the quality of the WikiDialog dataset. An illustrative example from the WikiDialog dataset is presented in Figure~\ref{fig:examples}. From this example, it is evident that combining certain answers could yield improved responses and questions. Moreover, some questions are overly broad, rendering them less suitable for training a retriever system. In the context of the Open-QA dataset, the "anything else" question serves as a means to transition between topics. However, when examining a brief paragraph, typically containing around six sentences or less, it becomes challenging to discern any significant shifts in the topic.


\section{\shortName}\label{sec:method}

\textbf{Problem Statement:} \shortName~(\textbf{Syn}thesizing \textbf{C}onversations using \textbf{A}utomatic \textbf{R}esponse \textbf{S}egmentation)~aims to generate a high-quality complete dialog from an informative document. It assumes that at most $N$ continuous sentences where ($N > 1$) from the document can be treated as an answer to an imagery question.

We build our work on top of the dialog inpainting~\cite{dai2022dialog}, where each sentence is treated as an answer (equivalent to $N=1$ in our assumption). Our idea stems from the observation that not all sentences are equally informative in WikiDialog, the dataset generated by dialog inpainting. Figure~\ref{fig:examples} shows some examples that demonstrate the limitations of the WikiDialog dataset.

To synthesize better ConvQA datasets, we implemented a simple sentence segmentation mechanism, along with a few modifications to the dialog inpainting method. In the following section, we will introduce these components in more detail.

\subsection{Notations}

Formally, a complete dialog $\mathbf{d}$ is a sequence of speaker questions, answers, and optional context, represented by:
\begin{equation}
    \mathbf{d} = (\mathbf{c} \oplus  \mathbf{q}_1\oplus \mathbf{a}_1\oplus \cdots \oplus \mathbf{q}_t \oplus \mathbf{a}_t \oplus \cdots),
    \label{eq:example}
\end{equation} where $\mathbf{q}_t$ and $\mathbf{a}_t$ are $t$ question and answer in a dialog, respectively. $\textbf{c}$ is the prefix optional context and $\oplus$ is the sequence joint symbol.  We use the same notation for partial dialogs, where unobserved questions are denoted by the $\cdot$ symbol. For example, $(\mathbf{c}\oplus \cdot\oplus \mathbf{a}_1\oplus \mathbf{q}_2\oplus \mathbf{a}_2\oplus \cdot\oplus \mathbf{a}_3)$ is a partial dialog where question $\mathbf{q}_1$ and $\mathbf{q}_3$ are unobserved, and we refer to these as "masked" utterances. Additionally, we use the shorthand $\mathbf{d}_{m(1,3)}$ to denote a dialog $\mathbf{d}$ with masked utterances at positions 1 and 3.

To complete the partial dialog $\mathbf{d}_{m(1,3)}$, we generate predictions for questions $1$ and $3$, denoted $\hat{q}_1$ and $\hat{q}_3$. The inpainted dialog is then:

\begin{equation}
        \scriptsize
        \text{Inpaint}(\mathbf{d}_{m(1,3)}) = (\textnormal{c}\oplus \hat{\mathbf{q}}_1\oplus \mathbf{a}_1\oplus \mathbf{q}_2\oplus \mathbf{a}_2\oplus \hat{\mathbf{q}}_3\oplus \mathbf{a}_3).
        \label{eq:inpaint}
\end{equation}

In this scenario, $\hat{\mathbf{q}}_1$ and $\hat{\mathbf{q}}_3$ are typically questions directed towards the next answer, and our goal is to associate them with all preceding utterances ($\mathbf{q}$ and $\mathbf{a}$).

\subsection{Answer Segmentation}
Each answer $\mathbf{a}_t$ in Eq.~(\ref{eq:inpaint}) can be further decomposed with sentences, denoted by:
\begin{equation}
    \mathbf{a}_t = (\mathbf{s}^1_t\oplus p^1_t\oplus \cdots\oplus \mathbf{s}^m_t\oplus p^m_t \cdots), 
\end{equation} where $\mathbf{s}^m_t$ is the $m$-th sentence in answer $\mathbf{a}_t$, and $p^m_t$ is its corresponding placeholder. Here we involve placeholder $p^m_t$ to aid in answer segmentation. Specifically, if $p^m_t$ is a special token (e.g., empty string in this paper), then we consider that $\mathbf{s}_{t}$ and $\mathbf{s}_{t+1}$ should be combined as one answer towards a question $\mathbf{q}_t$. Considering a similar case in Eq.~(\ref{eq:inpaint}), our inpainted dialog with answer segmentation can be written as follows:
\begin{equation}
    \scriptsize
    \text{SegInpaint}(\mathbf{d}_{m(1,3)}) = (\mathbf{c}\oplus \hat{\mathbf{q}}_1\oplus \mathbf{s}^1_1\oplus \hat{\mathbf{p}}^1_1\oplus \mathbf{s}^2_1\oplus \mathbf{q}_2\oplus \mathbf{s}^1_2\oplus \hat{\mathbf{q}}_3\oplus \mathbf{s}^1_3).
    \label{eq:seg}
\end{equation}

Our model is capable of generating questions ($\hat{\mathbf{q}}_1$ and $\hat{\mathbf{q}}_3$) and performing answer segmentation ($\hat{\mathbf{p}}^1_1$) simultaneously. If $\hat{\mathbf{p}}^1_1$ is the special token that we defined, then $\mathbf{q}_1$ is considered as the question to $(\mathbf{s}^1_1 \oplus \mathbf{s}^2_1)$. 
Otherwise, $(\hat{\mathbf{q}}_1\oplus\mathbf{s}^1_1)$ and $(\hat{\mathbf{p}}^1_1 \oplus \mathbf{s}^2_1)$ form two question-answer pairs. \textit{By combining some of those sentences, we can generate more comprehensive responses as well as improved questions.}

\subsection{Training}

To train our model, we utilize a partial dialog and aim to predict two values: $\mathbf{q}_t$ and $p^i_t$. This task is similar to the masked language modeling used in BERT~\cite{kenton2019bert}, where missing tokens in a passage are reconstructed. However, in our case, we aim to reconstruct a missing utterance in a dialog.

Let us assume that the model is a generative model with parameters $\mathbf{\theta}$, which specify a probability distribution $P_\theta(\mathbf{q}_t \mid \mathbf{d}_{m(t)})$. Our training objective is to minimize the standard cross-entropy loss:

\begin{equation}
    \mathcal{L}(\theta) = - \sum_{\mathbf{d}\in \mathcal{D}} \mathbb{E}_{\mathbf{q}_t \sim \mathbf{d}}\left[\text{log}\,P_{\mathbf{\theta}}(\mathbf{q}_t \mid \mathbf{d}_{m(t)}) \right]
\end{equation} where $\mathcal{D}$ is the set of complete dialogs and $\mathbf{q}_t$ is a randomly sampled question from the dialog $\mathcal{D}$.

\begin{figure}[htb]
  \centering
  \includegraphics[width=\linewidth]{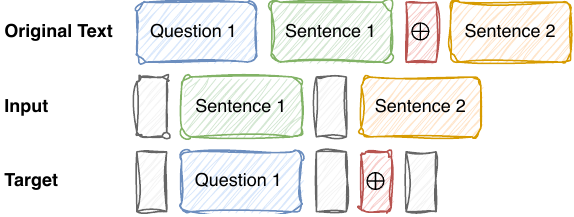}
  \caption{An illustration of preparing the training dataset, considering a training instance with a question and two sentences as answers. Here, grey boxes represent the extra\_ids tokens for T5. }\label{fig:training-examples}
\end{figure}

Following Dialog Inpainting~\cite{dai2022dialog}, we used T5~\cite{raffel2020exploring}, a text-to-text encoder-decoder Transformer as our pre-trained model. T5 uses a denoising objective that is slightly different from the original Masked Language Modeling (MLM) used in BERT. We believe that T5's denoising pre-training objective and encoder-decoder architecture are the most suitable for this task. Figure~\ref{fig:training-examples} shows the original texts, inputs and targets during our training.

During training, we randomly masked at least one and at most $N$ continuous questions $\mathbf{q}$ within a dialogue or question-answer pairs, as well as all answer segmentation placeholders $p$. As mentioned earlier in Section~\ref{sec:method}, our assumption is that $N$ is the maximum number of sentences that can form an answer. To balance the contextual awareness of the model, we decided to randomly add or remove titles of the dialogue or QA during training. 

\subsection{Inference}

\begin{figure}[htb]
  \centering
  \includegraphics[width=0.7\linewidth]{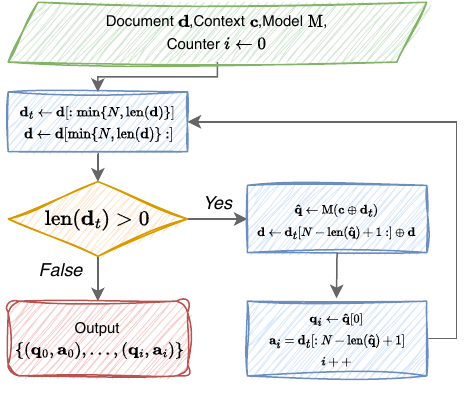}
  \caption{A flowchart illustrating the use of a trained model to convert a document into a dialogue format where, in each iteration, at most $N$ sentences are processed, and only the first generated question is retained.}\label{fig:inference}
\end{figure}

Figure~\ref{fig:inference} illustrates the process of using a trained model to convert a document into a dialogue format. The process begins with a document ($\mathbf{d}$) that contains a set of sentences. The trained model takes this document as input, along with a context ($\mathbf{c}$) that includes the document's title. The model then generates a dialogue by predicting questions and their corresponding answers based on the input sentences.

The model processes the document iteratively. In each iteration, it considers a fixed number of sentences (denoted as $N$ in the figure) and generates a question that summarizes these sentences. The generated question and its corresponding answer are then added to the dialogue history. This process continues until all the sentences in the document have been processed.

During the inference process, the model maintains a counter to keep track of the number of sentences processed so far. The output of the model is a dialogue consisting of generated questions and their associated answers, which together provide a comprehensive summary of the input document.

By following this approach, the trained model can effectively convert a document into a more accessible and interactive dialogue format, enabling users to quickly grasp the key points of the document through a series of relevant questions and answers.

\section{Experimental Setup}

\subsection{Datasets}

The successful implementation of answer segmentation highly depends on the training dataset. Naturally, if the training dataset only contains short answers or single sentences, the answer segmentation will fail. So we perform some basic data analysis in this section. Table~\ref{tab:sentences} shows the distribution of the number of sentences from the selected training datasets. Dialog Inpainting use OR-QuAC~\citep{qu2020open}, and QreCC~\citep{anantha2021qrecc} to conduct semi-supervised training, where more than 90\% of the answers consist of only one sentence. In order to improve the answer segmentation capacity, we utilized an additional Dolly dataset~\citep{databricks2023dolly}, where answers usually have more than one sentence. Based on the statistics of those datasets, we set $N = 3$ for our experiments reported in this paper.

\begin{table}[ht]
  \centering
  \scriptsize
  \begin{tabular}{lrrrrr}
    \toprule
    \textbf{Dataset}  & \textbf{Avg \# Sen.}  & \textbf{1 Sen.} & \textbf{2 Sen.} & \textbf{$\mathbf{\geq3}$ Sen.} \\
    \midrule 
     OR-QuAC & 1.08 & 92.52\% & 6.99\% & 0.49\%\\
    QreCC & 1.10 & 90.03\% & 9.29\% & 0.67\%\\
    Dolly & 3.44 & 43.18\% & 13.83\% & 42.99\% \\
    \bottomrule
  \end{tabular}
    \caption{The distribution of the number of sentences from the selected training datasets.}
  \label{tab:sentences}
\end{table}

We noticed that the WikiDialog dataset often includes a specific follow-up question -- "Are there any other interesting aspects about this article?". Figure~{\ref{fig:examples}} also shows this behavior. This follow-up question is a common sentence in the QreCC, and  OR-QuAC datasets. For instance, in QreCC, approximately 4\% of question-answer pairs and 22.9\% of dialogs contain this type of question. The original objective of those "anything else" questions was to indicate shifts in the current topic and to request any new information. However, generating these questions is not ideal because they lack the specificity needed to elicit answers that are meaningfully representative of the content being discussed. Simultaneously, we believe that a short documents should be within a topic. So we decided to cleanup QreCC and  OR-QuAC datasets, using a hand-crafted rule that excludes any question which contain "other interesting". \textbf{Thus, our synthetic data is less likely to generate questions regarding the shift of topics.}

Furthermore, each entry in QreCC and  OR-QuAC datasets contains two question types, the raw question (RQ) and the rewritten question (WQ). In cases where the raw question includes personal pronouns (such as "she," "they," and "we") and demonstrative pronouns (such as "these," "this," and "that"), a question rewriting model may involve hand-crafted rules to replace those pronouns or rewrite the question entirely. See~\citet{qu2020open, anantha2021qrecc} for more details. In this paper, we chose to use both question types during training as a data augmentation technique. Additionally, to control the desired output question types, we added a prefix to the input as "Type: \{\textit{question type}\}" to indicate the current question type for the placeholder. 

\subsection{Models}

In this study, we decided to use \textsc{FLAN-T5-XL}~\cite{chung2022scaling} due to the limitations in our computational resources. Although \textsc{T5-XXL}~\cite{raffel2020exploring}, an 11B parameter model, was used in dialog inpainting~\cite{dai2022dialog}, we opted for a smaller model. In Table~A.\ref{tab:ablation-gpt-eval} (provided in the Appendix), we show that there is not much difference in performance when applying the proposed method to either \textsc{FLAN-T5-XL} or its counterpart \textsc{T5-V1\_1-XL}. Both models are effective and efficient compared to the dialog inpainting approach.

To summarize, we initialized our model with \textsc{FLAN-T5-XL}, which has 3 Billion parameters, and fine-tuned it with 8 V100 16GB GPUs. The training process employed a constant learning rate of $10^{-4}$, a dropout rate of 0.1, an equivalent batch size of 32, and ran for 3.5K iterations (equivalent to 4 epochs).

\section{Evaluation}

Our primary focus in this study centers around conducting a human evaluation to compare the outputs generated by our approach with those produced by WikiDialog (WD) using Dialog inpainting method. To carry out this assessment, we utilize Amazon Mechanical Turk (MTurk) as our platform. Each human annotator is compensated at a rate of $0.036$ US dollars per question, and we provide them with identical instructions and sample examples as outlined in Figures 6 to 10 in the work of~\citet{dai2022dialog}. A potential issue in~\citet{dai2022dialog} lies in their use of the same raters for evaluations, which could introduce bias due to the evaluators' perception of the subject matter. To mitigate this concern, our evaluation process involves presenting subjective questions to a minimum of three distinct human evaluators for each dialogue turn. We label the answer for each question as the one agreed upon by at least two human annotators. Given that MTurk offers a more diverse pool of human evaluators, we use two-proportion z-test for evaluations. 

We report our findings based on dialogs that correspond to a set of 50 selected passages\footnote{These passages are the first 50 passages in the WD dataset and can be found at~\url{https://github.com/google-research/dialog-inpainting}}. Recent studies suggest using large language models (LLMs) as reference-free metrics for evaluating natural language generation (NLG) tasks. LLMs have the advantage of being applicable to new tasks that lack human-generated reference texts~\cite{liu-etal-2023-g}. Furthermore,~\citet{faysse2023revisiting} demonstrate that LLMs are more aligned with human preferences and exhibit consistent performance across a diverse set of generative tasks. In this study, we utilized the OpenAI chat completion API with the \textsc{GPT-4} model~\cite{brown2020language} to perform the same evaluation as human annotators. We used the exact same rubric as~\citet{dai2022dialog}. The rubrics and prompt templates used for both human and \textsc{GPT-4} evaluations are provided in Table~A.\ref{tab:prompt} in the Appendix. The results of this evaluation are presented in Table~\ref{tab:human-eval}.

\begin{table*}[tbh]
  \centering
  \begin{tabular}{lrr|r|rr|r}
    \toprule
    \textbf{Evaluator}& \multicolumn{3}{c|}{\textbf{Human}} & \multicolumn{3}{c}{\textbf{GPT 4}}\\
    \midrule
    \textbf{Answer}  & \textbf{RQ}  & \textbf{WD} & \textbf{WQ}   & \textbf{RQ} & \textbf{WD} & \textbf{WQ}  \\
     \midrule
     \multicolumn{1}{l}{\textit{Is the question information seeking?}}\\
     \textit{Yes (\%)}  &  82.6 &  \textbf{87.1} &  82.8 &  \textbf{99.3} &  99.0 &  100.0 \\
     \midrule
     \multicolumn{1}{l}{\textit{How relevant is question to the conversation?}}\\
     \textit{Not at all (\%)}  &   3.3 &   4.1 &   2.2 &  4.9 &  2.4 &   0.0 \\
     \textit{Topic only (\%)} &  45.0 &  45.5 &  43.0 &  20.3 &  26.4 &  22.0 \\
     \textit{Follows up (\%)} & \textbf{51.7} &  50.4 &  54.8 &  \textbf{74.8} &  71.2 &  78.0 \\
     \midrule
      \multicolumn{1}{l}{\textit{How speciﬁc is the question?}}\\
      \textit{Not at all (\%)} & 4.5 &   4.8 &   5.3 &  1.0 &  10.6 &  0.6 \\
      \textit{Somewhat (\%)} & 42.4 &  46.9 &  47.3 &  24.4 &  31.0 &  16.9 \\
      \textit{Very (\%)} & \textbf{53.1} &  48.3 &  47.4 &  \textbf{74.6} &  58.4 &  82.5 \\
      \midrule
      \multicolumn{1}{l}{\textit{How well answered is the question?}}\\
      \textit{Not at all (\%)} & 3.1 &   2.7 &   7.7 &   13.0 &   11.8 &   9.0 \\
      \textit{Incompletely (\%)} &  10.3 &  13.7 &   1.9 &  22.3 &  38.0 &  23.7 \\
      \textit{Sufﬁciently (\%)} &  40.9 &  40.4 &  47.4 &  21.7 &  21.6 &  19.3 \\
      \textit{Perfectly (\%)} &  \textbf{45.7} &  43.2 &  43.0 &  \textbf{43.0} &  28.6 &  48.0 \\
    \bottomrule
  \end{tabular}
  \caption{Results from a human evaluation of the generated dialog in four variants of our method vs. WikiDialog. In this evaluation, `RQ' represents the questions generated by our proposed method, `WQ' indicates rewritten questions, and `WD' represents questions generated by WikiDialog.  Our findings indicate that our proposed method's `RQ' outperforms WD in 7 out of the 8 cases.}\label{tab:human-eval}
\end{table*}

\subsection{Quality Comparison}

Overall, our approach consistently outperforms WikiDialog (WD) in terms of generating more specific questions and better answers, despite our model's smaller size. To assess the statistical significance of these improvements, we conducted a two-proportion z-test, which is a statistical test used to determine if the proportions of categories in two group variables significantly differ from each other. This means that it is suitable when your variable of interest is categorical and have more than 10 values in each of the populations. If we consider `Very' as an acceptable answer for the question `How specific is the question?', then RQ is significantly better than WD with a p-value of $2.5\times10^{-2}$ according to human assessment and $3.0\times10^{-4}$ according to \textsc{GPT-4}. This indicates that our proposed method excels in asking more specific questions compared to WD. 

We also conducted a similar test on the criterion `How well answered is the question?' with `Perfectly' as an acceptable answer choice. In this evaluation, RQ once again outperforms WD, with a p-value of $4.7\times10^{-4}$ according to human assessment and $2.0\times10^{-3}$ according to \textsc{GPT-4}. This demonstrates that RQ achieves better answer quality than WD. The improved question specificity and answer quality can enhance the utility of the synthesized dataset for downstream tasks, such as information retrieval. While WQ and RQ are both superior to WD in terms of question relevance based on human and \textsc{GPT-4} evaluations, these differences are not statistically significant. In summary, RQ outperforms WD in 7 out of the 8 cases, demonstrating the superiority of our proposed method compared to the baseline.  

\subsection{Question Types}

Unlike the WikiDialog dataset, we offer rewritten questions (WQ). The expectation is that these rewritten questions are superior based on many proposed criteria, especially since they tend to contain fewer personal and demonstrative pronouns. This assumption is also validated in Table~\ref{tab:human-eval}. However, for downstream tasks or real-world scenarios where users provide natural inputs, questions are less often in this rewritten style.

\subsection{GPT-4 vs Human Evaluation}

In general, from our comparison analysis results, GPT-4 evaluation is aligned with our human evaluation in most of the cases, which supports the argument that GPT-4 evaluation is helpful~\cite{liu-etal-2023-g,faysse2023revisiting}.

We have also observed that GPT-4 become more "binary thinking" than our human evaluators. GPT-4 tends to output the highest ordinal variable while human evaluators seems more conservative. Those evidences could be found from Table~\ref{tab:human-eval} that GPT-4 has higher absolute difference between largest ordinal variable and the second largest one compared to the one produced by human.

There is a discrepancy, though no statistical differences, between \textsc{GPT-4} and humans in determining whether a question is information-seeking or not. The discrepancy can be attributed to the presence of `anything else' questions in WD. When GPT-4 assesses `anything else' questions, it categorizes them as information-seeking with 100\%, while other questions have a 98.2\% chance of being considered as information-seeking. In contrast, humans treat 79.6\% of `anything else' questions as information-seeking, while 88.0\% of the remaining questions are also considered information-seeking. Additionally, the disagreement may arise from the fact that `anything else' questions often lead to more vague or general answers in the training data for \textsc{GPT-4} causing it to classify such questions as less information-seeking.

\section{Application to Conversational Retrieval}

As mentioned earlier in the introduction section, there are few ways to utilize the generated dataset, for e.g.,  use the dataset to train chatbot directly or transform it as a open-domain conversation retrieval task. In this section, we will focus on open-domain conversational retrieval, as datasets such as OR-QuAC~\cite{qu2020open}, Trec-CAsT-19~\cite{dalton2020trec}, and Trec-CAsT-20~\cite{dalton2021cast} exist in this domain.

A ConvQA system interacts with a user in a multi-turn dialogue, where the user primarily asks questions and the system responds (with occasional exceptions, such as the system asking for clarification). When it is the system's turn to speak at a given time $t$, it considers the entire dialogue history, comprising all previous turns, and generates a new utterance as its response. This work focuses on the conversational retriever, showing how to improve it by pre-training on our dataset comparing to WD, leaving improvements to the generator for future work.

\subsection{Dual Encoder}

In our methodology, we employ a standard dual encoder as described in \citet{ni2022sentence}. The objectives involve optimizing for a combination of factors: maximizing the similarity between a query $\mathbf{q}$ and its corresponding positive passage $\mathbf{p}^*$, while simultaneously minimizing the similarity between query $q$ and all of its negative samples $\mathcal{N}(\mathbf{p})$. This is achieved through the following loss function:

\begin{equation}
l(\boldsymbol{\theta}) = - \log \frac{\exp( s_{\boldsymbol{\theta}} (\mathbf{q}, \mathbf{p})/\tau)}{\sum_{\mathbf{p} \in {\mathbf{p^*}\cup\mathcal{N}(\mathbf{p})}} \exp(s_{\boldsymbol{\theta}}(\mathbf{q}, \mathbf{p})/\tau )}
\end{equation}

Here, $s_{\boldsymbol{\theta}}(\mathbf{q}, \mathbf{p})$ represents a standard cosine similarity, defined as:

\begin{equation}
s_{\boldsymbol{\theta}}(\mathbf{q}, \mathbf{p}) = \frac{\text{embed}_{\boldsymbol{\theta}} (\mathbf{q})^\intercal \text{embed}_{\boldsymbol{\theta}}(\mathbf{p})}{\lVert\text{embed}_{\boldsymbol{\theta}} (\mathbf{q}) \rVert \lVert\text{embed}_{\boldsymbol{\theta}} (\mathbf{p})\rVert }
\end{equation}

In this context, $\text{embed}_{\boldsymbol{\theta}}$ refers to an embedding model used to map text into a fixed-dimension embedding vector. We adopt the setup outlined in \citet{dai2022dialog}, utilizing a pre-trained \textsc{T5-Large} encoder as the $\text{embed}$ model.

\subsection{Datasets}

The entire WD dataset consists of a total of 11.3 million dialogs. This comprehensive dataset is divided into 100 separate sections, commonly referred to as "dumps." However, due to the sheer size of the WD dataset, utilizing all of it for demonstration purposes can be overwhelming. Therefore, to simplify the demonstration and still provide meaningful insights, we are focusing only on the first five dumps, labeled as \#00000 through \#00004. These chosen dumps are used to create the RQ and WQ datasets. Notably, each dump from the RQ and WQ datasets represents 1\% of the entire WD dataset. For readers interested in a more detailed statistical comparison between our generated datasets (RQ and WQ) and the original WD dataset, please refer to Appendix Table~A.\ref{tab:ablation:stats}.


\subsection{Two-Stage Training Strategy}

\begin{figure}[htb]
  \centering
  \includegraphics[width=\linewidth]{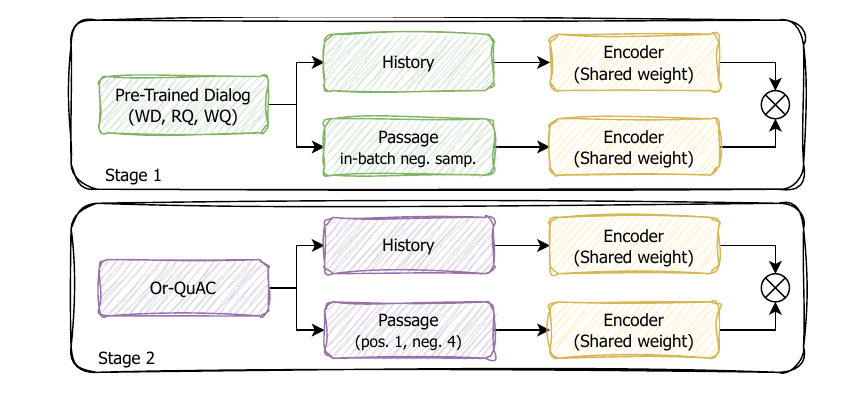}
  \caption{Our dual encoder employs a two-stage training approach. $\otimes$ represents the cosine similarity. Initially, the T5-encoder is pre-trained using the generated dataset (WD, RQ, WQ). Subsequently, it is fine-tuned using the OR-QuAC dataset.  }\label{fig:dual-encoder}
\end{figure}

We employ a two-stage training approach for our dual encoder. Figure~\ref{fig:dual-encoder} provides more details of our training strategy.

In the first phase, we use our generated dataset, comprising WD, RQ, and WQ, to train $\text{embed}_\theta$. This is accomplished by implementing in-batch negative sampling, whereby the positive passage for a given instance, represented by $i$, is treated as a negative sample for all other instances in the same batch that do not equal  to $i$. This in-batch negative sampling strategy is an efficient, straightforward method to train our sentence embedding model. To summarize our initial training phase, we used the aforementioned datasets with a mini-batch size of 8, a learning rate of $1 \times 10^{-4}$, an iteration step of 500 for each 1\% subset, and the accumulated gradient batches parameter set to 32 and AdamW optimizer~\cite{loshchilov2017decoupled}.

In the second phase, we fine-tune the model using  OR-QuAC training set. For simplicity and in the interest of fair comparison, we've chosen not to incorporate the mutli stage hard sampling method outlined in~\citet{dai2022dialog}. Instead, we only use the annotated positive sample and negative samples in the dataset. Each model employed here was trained using the following parameters: a mini-batch size of 16, a learning rate of $1 \times 10^{-4}$, an iteration step of 250, and the AdamW optimizer for weight adjustments. These parameters were chosen to ensure the robustness of our models.

\subsection{Results}

\begin{table}[th]
  \centering
  \scriptsize
  \begin{tabular}{lrrr}
    \toprule
    \textbf{System} & \textbf{OR-QuAC}&\textbf{CAsT-19}&\textbf{CAsT-20}\\
    & \textbf{MRR@5} & \textbf{MRR} & \textbf{MRR}\\
    \midrule
    BM25-QR & 20.2 & 58.1 & 25.0\\
    ANCE-QR & 45.7 &  66.5 & 37.5\\ 
    ConvDR & 61.6 & 74.0 & 50.1\\ 
    \midrule
    T5-Large DE & 57.3 & 61.1 & 34.5\\
    \;\;\; + 1\% WD & 60.1 (0.53) &62.6 (0.41) &37.2 (0.85)\\
    \;\;\; + 1\% RQ  & 60.9 (0.71) &63.5 (0.92)&37.4 (0.40)\\
    \;\;\; + 1\% RQ + 1\% WQ & 62.1 (0.18)&64.3 (0.69)&38.2 (0.34)\\
    \;\;\; + 5\% RQ + 5\% WQ & 62.8&64.7&38.5\\
    \bottomrule
  \end{tabular}
    \caption{Performance analysis of a Retrieval-Only Dual Encoder on the OR-QuAC, Trec-CAsT-19 and Trec-CAsT-20. Key abbreviations include DE for Dual Encoder, WD for WikiDialog Dataset, RQ for generated questions, and WQ for rewritten questions. For subsets labeled with `1\%', results from five distinct experiments are averaged and presented using mean (standard deviation). The Retrieval-Only Dual Encoder, fine-tuned on our generated datasets RQ and WQ, demonstrates superior performance compared to the Retrieval-Only Dual Encoder fine-tuned on the original WD dataset and other existing baselines.}
  \label{tab:retrivel}
\end{table}

During our experimentation, we discovered that using a small portion of the dataset for training can still improve the performance of the dual encoder. This finding offers valuable insights for optimizing computational efficiency without substantially compromising the effectiveness of the model. For example, in our cases, finetuning on $1\%$ of RQ can lead to an average $0.8\%$ performance improvement comparing to using the same WD for OR-QuAC dataset. Upon analysis, our generated datasets (RQ and WQ) consistently exhibited superior performance when compared to the original WD dataset in the context of open-domain conversational retrieval tasks, and this superiority is statistically significant.

In addition, we evaluated our pre-trained dual-encoder retrievers in comparison to three well-known retrieval-only benchmarks:  BM25-T5QR by~\citet{wu2022conqrr}, ANCE-Query Rewriter by~\citet{yu2020few} and ConvDR by~\citet{yu2021few}. Our approach outperforms all of these existing baselines. \textit{This conclusion not only confirms the quality of our data generation methods but also highlights their potential applicability and utility in enhancing the model's ability to address the inherent complexities of such tasks.} Additionally, as we increase the size of the fine-tuning dataset from 1\% to 5\% percent, the performance improves further.  

\section{Conclusion}

In this paper, we introduce~\shortName, a novel approach that generates high-quality synthetic conversations from unlabeled documents by leveraging ConvQA datasets. Our method outperforms existing benchmarks in terms of the quality of the generated conversations. To generate high-quality synthetic data, we developed an answer segmentation technique that incorporates a special token and curated a new dataset. Despite using a smaller model and lower compute, human evaluations show that our generated dataset surpasses the WikiDialog dataset in terms of answer quality and question specificity. This demonstrates the effectiveness of the proposed approach. Moreover, the question retrieval system trained on our dataset outperforms both the standard retrieval-only benchmark and the same model trained using the WikiDialog dataset. We believe our contributions will facilitate future progress in the development of document-based conversational systems.

\section*{Limitations}

In our computational setup, we employed machines that were equipped with 8 $\times$ V100 GPUs to train the \textsc{FLAN-T5-XL} model. This training process involved using a mixed dataset comprising three public datasets. To give an estimate of the time required, training the \textsc{FLAN-T5-XL} model typically takes approximately one day.  These time estimates highlight the significant computational resources needed for training both models effectively.

For the retrieval experiments, it is important to note that the performance of the in-batch negative sampling strategy is significantly influenced by the mini-batch size. As the size of the mini-batch increases, model performance typically improves, since each iteration introduces more negative samples. However, due to constraints on the computational resources, we had to keep the mini-batch size relatively small.

In addition, we are unable to replicate the experiment conducted by~\citet{dai2022dialog} and train our model using  a full dataset (either 100\% WD or 100\% RQ and WQ) due to our limited computing resources. It is important to note that this limitation may impact the validation of the model's effectiveness. The absence of evidence in this regard leaves uncertainty about the model's performance in this context.

\section*{Ethics Statement}

This paper does not present any ethics-related issues. The data and additional resources utilized in this work are open-source and widely used in existing research.

{\small\bibliography{reference.bib}}

\onecolumn

\appendix

\section*{Appendix}

\captionsetup[table]{labelformat=AppendixTables}
\captionsetup[figure]{labelformat=AppendixFigures}
\setcounter{table}{0}
\setcounter{figure}{0}

\subsection*{A.1 Choice of the base model}
We conducted an ablation study to check which base model to use. Table~A.\ref{tab:ablation-gpt-eval} shows the result for \textsc{GPT-4} evaluation. Using either \textsc{T5-XL} or \textsc{FLAN-T5-XL}, our method produces successful results in the \textsc{GPT-4} evaluation. We also added LLAMA-7B based on zero-shot setting as baseline method for reference.

\begin{table}[ht]
  \centering
  \scriptsize
  \begin{tabular}{lp{0.05\linewidth}|p{0.05\linewidth}p{0.05\linewidth}|p{0.05\linewidth}p{0.05\linewidth}|p{0.05\linewidth}p{0.05\linewidth}}
    \toprule
    \textbf{Model } & \multicolumn{1}{l|}{\textbf{T5-XXL}} & \multicolumn{2}{l|}{\textbf{T5-XL}}& \multicolumn{2}{l}{\textbf{FLAN-T5-XL}}&\multicolumn{1}{l}{\textbf{LLAMA2-7B}}\\
    \cmidrule{2-7}
     \textbf{Dataset} & \textbf{WD} & \textbf{RQ} & \textbf{WQ} & \textbf{WD} & \textbf{RQ} & \textbf{0-shot} \\
     \midrule
     \multicolumn{1}{l}{\textit{Is the question information seeking?}}\\
     \textit{Yes (\%)} &99.0&100.0&100.0&99.3&100.0&88.8\\
     \midrule
     \multicolumn{1}{l}{\textit{How relevant is question to the conversation?}}\\
     \textit{Not at all (\%)} &2.4&1.5&0.0&4.9&0.0&4.0\\
     \textit{Topic only (\%)} &26.3&15.0&22.1&20.3&22.0&23.7\\
     \textit{Follows up (\%)} &71.2&83.5&77.9&74.8&78.0&71.8\\
     \midrule
      \multicolumn{1}{l}{\textit{How speciﬁc is the question?}}\\
      \textit{Not at all (\%)} &10.6&1.1&1.2&1.0&0.6&12.0\\
      \textit{Somewhat (\%)}  &31.0&30.1&19.1&24.4&16.9&35.6\\
      \textit{Very (\%)}  &58.4&68.9&79.7&74.6&82.5&52.0\\
      \midrule
      \multicolumn{1}{l}{\textit{How well answered is the question?}}\\
      \textit{Not at all (\%)} &11.8&11.5&9.9&13.0&9.0&17.3\\
      \textit{Incompletely (\%)} &38.0&16.9&18.0&22.3&23.7&50.3\\
      \textit{Sufﬁciently (\%)}  &21.6&25.2&22.7&21.7&19.3&20.8\\
      \textit{Perfectly (\%)} &28.6&46.4&49.4&43.0&48.0&11.6\\
    \bottomrule
  \end{tabular}
  \caption{Results from a \textsc{GPT-4} evaluation of 50 generated dialogs in four variants of our method vs. WikiDialog. Here RQ are generated questions, WQ is rewritten questions and WD is WikiDialog.}\label{tab:ablation-gpt-eval}
\end{table}

\subsection*{A.2 Statistics of the generated dataset}

The WikiDialog (WD) dataset consists of a total of 11.3 million dialogues. This dataset is divided into 100 separate partitions for easier handling and processing. We use the first five partitions (\#00000 to \#00004) to generate new dialogs and label them as RQ and WQ datasets. Appendix Table~A.\ref{tab:ablation:stats} shows the statistics of our generated dataset compared to the WD dataset. While there are a few entries for which our model cannot generate outputs in the correct format, the number of dialogs in the RQ and WQ datasets for each partition is slightly fewer than that in the WD dataset. The dialogues in our dataset generally consist of a small number of turns, which aligns with our objective of combining multiple responses to create a single, comprehensive answer.

\begin{table}[ht]
  \centering
  \scriptsize
  \begin{tabular}{lrrrrr}
    \toprule
    \textbf{Stat.} &\textbf{Part. \#} & \textbf{WD} & \textbf{RQ}& \textbf{WQ} \\
    \midrule
    \textit{\# Dialog} & 00000 &113,678&113,650&113,609 \\
    &00001 & 113,651 &113,613&113,574  \\ 
    &00002 & 113,536 &113,498&113,466&  \\ 
    &00003 & 114,286 &114,263& 114,221 \\ 
    &00004 & 113,596 &113,571&113,542  \\ 
    \midrule
    \textit{Avg. Turn} & 00000 &4.93&3.55&3.49 \\
    &00001 & 4.93 &3.57& 3.50  \\ 
    &00002 & 4.93 &3.56&3.49&  \\ 
    &00003 & 4.93 &3.56&3.49& \\ 
    &00004 & 4.93 &3.56&3.50 \\ 
    \bottomrule
  \end{tabular}
    \caption{Statistics of WD, RQ, and WQ datasets. Here, RQ is the dataset with the generated raw questions, WQ is the dataset with rewritten questions and WD corresponds to the WikiDialog dataset.}
  \label{tab:ablation:stats}
\end{table}

\newpage
\subsection*{A.3 Question and Answer Overlapping}
Table~A.\ref{tab:ablation:overlapping} shows the overlap between questions and answers using the ROUGE score. We can conclude that our RQ and WQ datasets have higher Rouge scores compared to WD dataset, although the absolute ROUGE scores still indicate a low level of text overlap. This further demonstrates the superior quality of our RQ and WQ datasets.

\begin{table}[ht]
  \centering
  \scriptsize
  \begin{tabular}{lrrrrr}
    \toprule
    \textbf{Dataset} &\textbf{ROUGE-1} & \textbf{ROUGE-2} & \textbf{ROUGE-L} \\
    \midrule
    \textbf{WD} & 0.111 & 0.026 & 0.095\\
    \textbf{RQ} & 0.158 & 0.046 & 0.130\\
    \textbf{WQ} & 0.205 & 0.080 & 0.166\\
    \bottomrule
  \end{tabular}
    \caption{ROUGE score for the generated question and the corresponding answers. }
  \label{tab:ablation:overlapping}
\end{table}

\newpage
\subsubsection*{A.4 Prompt Template for \textsc{GPT-4} evaluation}
In Table~A.\ref{tab:prompt}, we show the prompt Template for \textsc{GPT-4} evaluation. The same rubric is also used for human evaluation.

\begin{table*}[ht]
    \centering
    \scriptsize
    \begin{tabular}{p{0.9\linewidth}}
    \toprule
    \textbf{Question:} \textit{Is the question information seeking?} \\
    \textbf{Prompt Template and Rubric:} \\
    Is the QUERY information-seeking based on RUBRIC?  Output option only\\
    option: \\
    * Yes \\
    * No \\
    RUBRIC: \\
    * Yes. The user is looking to learn some information from the system. Note: Information-seeking queries don't have to be phrased as questions. \\
    * No. The query is unclear, difficult to understand or not seeking information. Note: Not all questions are information seeking, e.g. questions directed at the system ("how are you", "what do you think") or ones that are nonsensical in the context ("Brian, how is Jill doing?"). \\
    CONVERSATION: \{conversation\} \\
    QUERY: \{\textit{query}\} \\
    ANSWER: \{\textit{answer}\} \\
    \midrule
    \textbf{Question:} \textit{How relevant is question to the conversation?} \\
    \textbf{Prompt Template and Rubric:} \\
    How is the QUERY relevant to a CONVERSATION based on RUBRIC? Output option only.\\
    option \\
    * A \\
    * B \\
    * C \\
    RUBRIC: \\
    * A. Follows up on a previous query or response. It is difficult to correctly understand the query without reading the conversation history.\\
    * B. It is difficult to correctly understand the query without reading the conversation history. Only related to the topic of the conversation. The query is topically similar to previous queries or responses, but can be understood without reading them.\\
    * C. Not relevant. The query doesn't appear to be relevant to the topic or a previous query or response. Rule of thumb: if you are surprised by a query, it is probably not relevant.\\
    CONVERSATION: \{conversation\} \\
    QUERY: \{\textit{query}\} \\
    ANSWER: \{\textit{answer}\} \\
    \midrule
        \textbf{Question:} \textit{How speciﬁc is the question?} \\
    \textbf{Prompt Template and Rubric:} \\
    How specific is the QUERY based on RUBIC? CONVERSATION is the history context. Only output option text.\\
    option \\
    * Very \\
    * Somewhat \\
    * Not at all \\
    RUBRIC: \\
    * Very. Only a specific answer would satisfy the user. Example: "Why did she make the news in 1999?" likely requires a very specific answer. \\
    * Somewhat. A variety of answers of a specific kind would satisfy the user. Example: While there are many possible answers to "What else does she do?", they are all likely to be a job or activity. \\
    * Not at all. Many topically different answers would satisfy the user. Example: "Tell me something interesting about her." can be answered in many different ways. \\
    CONVERSATION: \{conversation\} \\
    QUERY: \{\textit{query}\} \\
    ANSWER: \{\textit{answer}\} \\
    \midrule
            \textbf{Question:} \textit{How speciﬁc is the question?} \\
    \textbf{Prompt Template and Rubric:} \\
    How well does the response ANSWER the QUERY based on RUBRIC? CONVERSATION is history context. Only output option text.\\
    option:
    * Perfectly \\
    * Sufficiently \\
    * Incompletely \\
    * Not at all \\
    RUBRIC: \\
    * Perfectly. The response completely satisfies the user's information need. \\
    * Sufficiently. The response mostly answers the user's information need, though some additional information could be provided. \\
    * Incompletely. The response provides some information relevant to the user, but doesn't adequately answer the question. \\
    * Not at all. The response does not provide any relevant information for the user's query or is not intelligible. \\
    CONVERSATION: \{conversation\} \\
    QUERY: \{\textit{query}\} \\
    ANSWER: \{\textit{answer}\} \\
    \hline
    \end{tabular}
    \caption{Prompt Template for \textsc{GPT-4} evaluation.}
    \label{tab:prompt}
\end{table*}

\clearpage
\subsubsection*{A.4 Additional Generated Examples}
\begin{figure*}[ht]
  \centering
  \includegraphics[width=\linewidth]{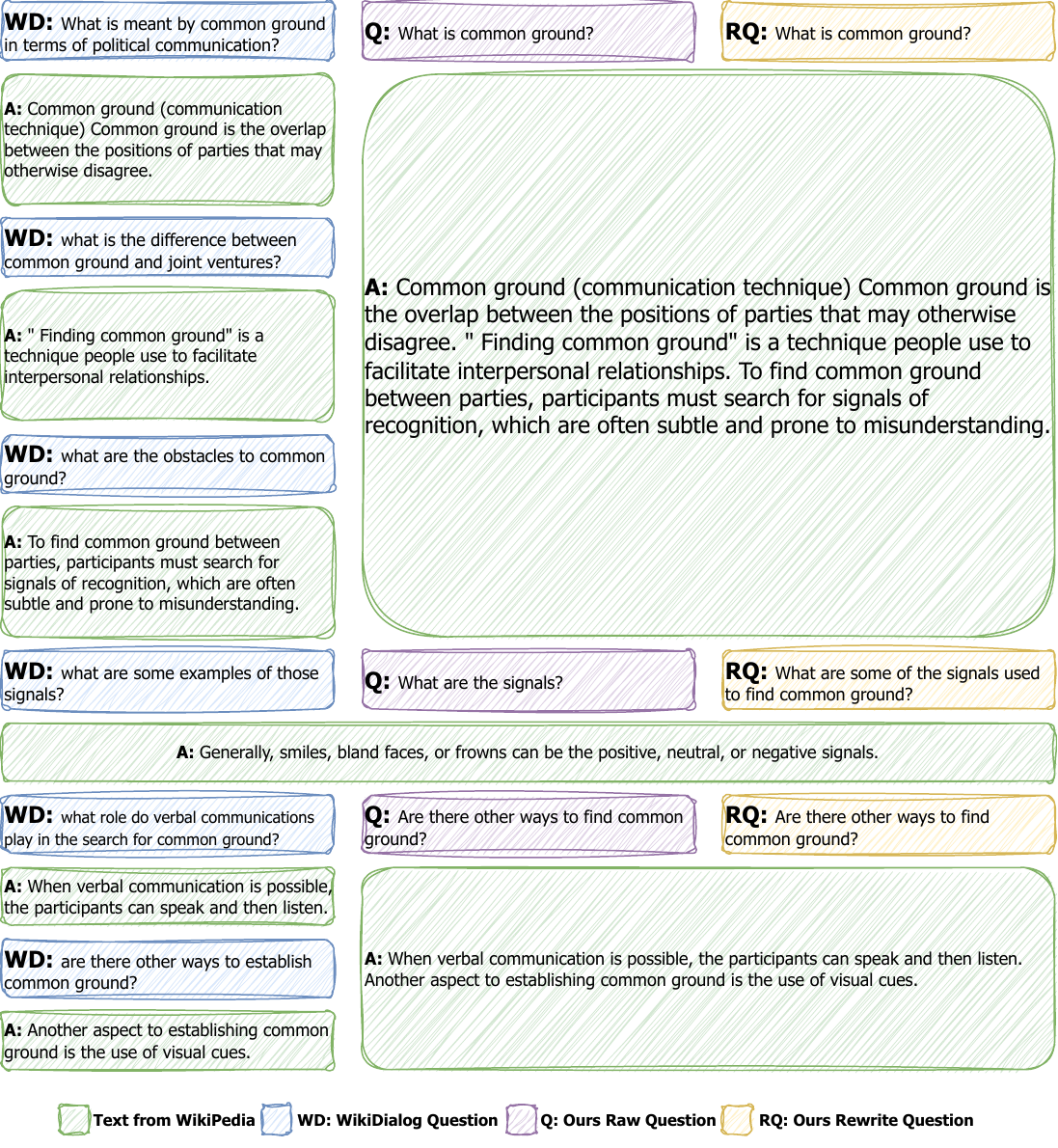}
  \caption{Comparison examples \#1. Here we can find the ability of our proposed method to perform segmentation of the sentences. This link (\url{https://en.wikipedia.org/wiki/Grounding_in_communication}) gives the raw Wikipedia web page.}\label{fig:examples-2}
\end{figure*}

\begin{figure*}[ht]
  \centering
  \includegraphics[width=\linewidth]{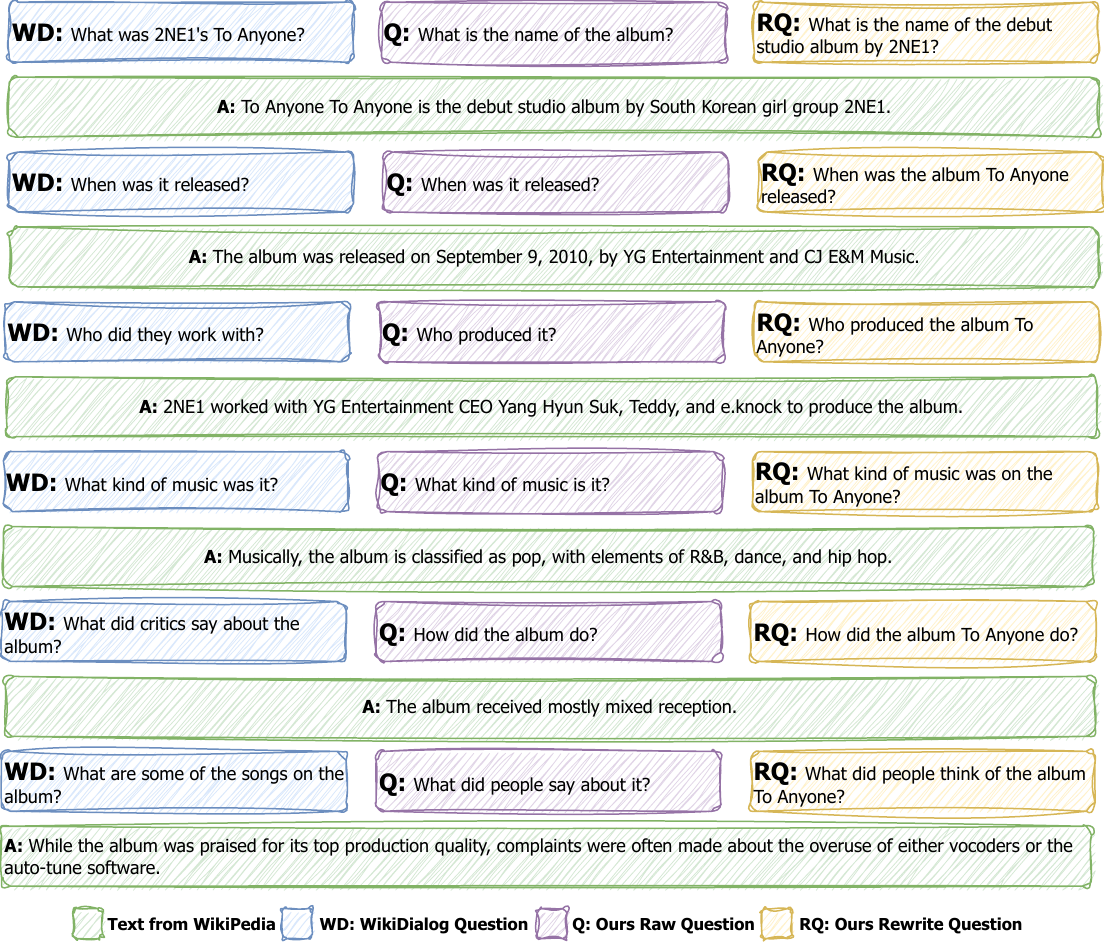}
  \caption{Comparison example \#2. Here we can find a relatively better question generation compared to WikiDialog. This link (\url{https://en.wikipedia.org/wiki/2NE1}) gives the raw Wikipedia web page.}\label{fig:examples-3}
\end{figure*}

In Figure~A.\ref{fig:examples-2} and Figure~A.\ref{fig:examples-3}, we show two more examples to compare performance with WD. 

\end{document}